\documentclass{article}
\usepackage{ijcai21}

\usepackage{times}

\usepackage{soul}
\usepackage{url}
\usepackage[hidelinks]{hyperref}
\usepackage[utf8]{inputenc}
\usepackage[small]{caption}
\usepackage{graphicx}
\usepackage{amsmath}
\usepackage{booktabs}
\usepackage[justification=centering]{caption}
\usepackage{footnote}
\title{Gated Transformer Networks for Multivariate Time Series Classification}

\author{
Minghao Liu$^1$\and
Shengqi Ren$^1$\and
Siyuan Ma$^{1}$\and
Jiahui Jiao$^1$\and
Yizhou Chen$^2$\and
Zhiguang Wang$^{3}$\And
Wei Song$^{1}$\footnote{Contact Author:Wei Song.\\
Email address:iewsong@zzu.edu.cn}\\
\affiliations
$^1$Zhengzhou University\\
$^2$Statistical information center of Henan health commission\\
$^3$Facebook AI\\

\emails
\{liuminghao, 202012172013047, 202012172013065, 202012172013044\}@gs.zzu.edu.cn,
xnhcyz@163.com,
zgwang813@gmail.com,
iewsong@zzu.edu.cn
}

\begin{document}

\maketitle

\begin{abstract}
Deep learning model (primarily convolutional networks and LSTM) for time series classification has been studied broadly by the community with the wide applications in different domains like healthcare, finance, industrial engineering and IoT. Meanwhile, Transformer Networks recently achieved frontier performance on various natural language processing and computer vision tasks. In this work, we explored a simple extension of the current Transformer Networks with gating, named Gated Transformer Networks (GTN) for the multivariate time series classification problem. With the gating that merges two towers of Transformer which model the channel-wise and step-wise correlations respectively, we show how GTN is naturally and effectively suitable for the multivariate time series classification task. We conduct comprehensive experiments on thirteen dataset with full ablation study. Our results show that GTN is able to achieve competing results with current state-of-the-art deep learning models. We also explored the attention map for the natural interpretability of GTN on time series modeling. Our preliminary results provide a strong baseline for the Transformer Networks on multivariate time series classification task and grounds the foundation for future research.
\end{abstract}

\section{Introduction}
We are surrounded by the time series data such as physiological data in healthcare, financial records or various signals captured the sensors. Unlike univariate time series, multivariate time series has much richer information correlated in different channels at each time step. The classification task on univariate time series has been studied comprehensively by the community whereas multivariate time series classification has shown great potential in the real world applications. 
Learning representations and classifying multivariate time series are still attracting more and more attention.

As some of the most traditional baseline,
distance-based methods work directly on raw time series
with some pre-defined similarity measures such as Euclidean distance and Dynamic Time Warping (DTW)\cite{2002Exact} to perform classification. With k-nearest neighbors as the classifier, DTW is known to be a very efficient approach
as a golden standard for decades. Following the scheme of well designed feature, distance metric and classifiers, the community proposed a lot of time series classification methods based on different kinds of feature space like distance, shapelet and recurrence, etc. These approaches kept pushing the performance and advanced the research of this field.

With the recent success of deep learning approaches on different tasks on the temporal data like speech, video and natural language, learning the representation from scratch to classify time series has been attracted more and more studies. For example, \cite{zheng2016exploiting} proposed a multi-scale convolutional networks for univariate time series classification. The author combines well-designed data augmentation and engineering approach like down sampling, skip sampling and sliding windows to preprocess the data for the multiscale settings, though the heavy preprocessing efforts and a large set of hyperparameters make
it complicated and the proposed window slicing method for data augmentation is not scalable. \cite{7966039} firstly proposed two simple but efficient end-to-end models based on convolutions and achieved the state-of-the-art performance for univariate time series classification. Thereafter, convolution based models demonstrate superior performance on time series classification tasks \cite{IsmailFawaz2018deep}.

Inspired by the recent success of the Transformer networks on NLP \cite{vaswani2017attention,devlin2018bert}, we proposed a transformer based approach for multivariate  time series classification. By simply scaling the traditional Transformer model by the gating that merges two towers, which model the channel-wise and step-wise correlations respectively, we show that how the proposed Gated Transformer Networks (GTN) is naturally and effectively suitable for the multivariate time series classification task. Specifically, our contributions are the following:

\begin{itemize}
\item  We explored an extension of current Transformer networks with gating, named Gated Transformer Networks for the multivariate time series classification problem. By exploiting the strength where the Transformers processes data in parallel with self-attention mechanisms to model the dependencies in the sequence, we showed the gating that merges two towers of Transformer Networks that model the channel-wise and step-wise correlations is very effective for time series classification task.
\item  We evaluated GTN on the thirteen multivariate time series benchmark datasets and compared with other state-of-the-art deep learning models with comprehensive ablation studies. The experiments showed that GTN achieves competing performance.
\item  We qualitatively studied the feature learned by the model by visualization to demonstrate the quality of the feature extraction of GTN.
\item  We preliminary explored the interpretability of the attention map of GTN on time series modeling to study how self-attention helps on channel-wise and step-wise feature extraction.
\end{itemize}

\section{Related Work}
The earlier work like \cite{yang2015deep} preliminary explored the deep convolutional networks on multivariate time series for human activity recognition. \cite{7966039} studied Fully Convolutional Networks (FCN) and Residual Networks (ResNet) and achieved the state-of-the-art performance for univariate time series classification. This work also explored the interpretability of FCN and ResNet with Class Activation Map (CAM) to highlight the significant time steps on the raw time series for a specific class. \cite{serra2018towards} proposed the Encoder model, which is a hybrid deep convolutional networks whose architecture is inspired by FCN with a main difference where the GAP layer is replaced with an attention layer to fuse the feature maps. The attention weight from the last layer is used to learn which parts of the time series (in the time domain) are important for a certain classification. \cite{DBLP:journals/corr/abs-1709-05206} proposed the two towers with Long short-term memory (LSTM) and FCN. The authors merge the feature by simple concatenation at the last layer to improve the classification performance on univariate time series, though LSTM bears the high computation complexity. Other earlier works like \cite{cui2016multi,zheng2016exploiting,le2016data,zhao2017convolutional} explored different convolutional networks architecture other than FCN and ResNet and claimed superior results on univariate or multivariate time series classifications, and served as the strong baselines in our work. Note that  \cite{tanisaro2016time} proposed Time Warping Invariant Echo State Network as one of the non-convolutional recurrent architectures for time series forecasting and classification, where we also include it in the study as a baseline.

There are plenty of established works on Transformer Networks for NLP and computer vision. The time series community is also exploring Transformers on forecasting and regression task, like \cite{li2019enhancing,cai2020traffic}. The studies on Transformer for time series classification is still in the early stage, like \cite{oh2018learning} explores the Transformer on clinical time series classification. The most recent work we discovered is from \cite{RUWURM2020421}, where the author studied the Transformer for raw optical satellite time series classification and obtained the latest results comparing with convolution based solutions. Our work gaps the bridge as the first comprehensive study for Transformer networks on multivariate time series classification.

\section{Gated Transformer Networks}

\begin{figure}
    \centering
    \includegraphics[scale=0.33]{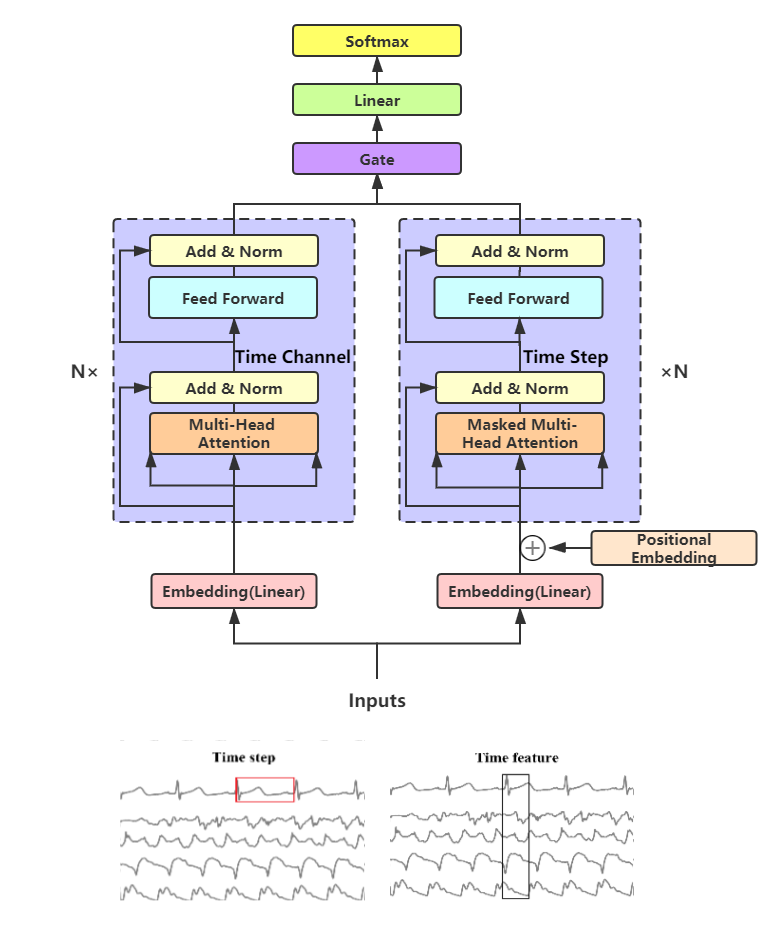}
    \captionsetup{justification=centering}
    \caption{Model architecture of the Gated Transformer Networks.}
    \label{fig:GTN}
\end{figure}

Traditional Transformer has encoder and decoder stacking on the word and positional embedding for sequence generation and forecasting task. As for multivariate time series classification, we have three extension simply to adapt the Transformer for our need - embedding, two towers and gating. The overall architecture of Gated Transformer Networks is shown in Figure \ref{fig:GTN}.

\begin{savenotes}
\begin{table*}[htbp]      
\centering
\caption{Test accuracy of GTN and other benchmark models on 13 multivariate time series dataset}
\label{results}
    \begin{tabular}{cccccccccc|c}    
        \toprule            
        & MLP & FCN & ResNet & Encoder & MCNN & t-LeNet & MCDCNN & Time-CNN & TWIESN	& GTN\\     
        \midrule            
        AUSLAN & 93.3& \textbf{97.5}& 97.4& 93.8& 1.1& 1.1& 85.4& 72.6& 72.4 & \textbf{97.5}\\ 
        ArabicDigits & 96.9& 99.4& \textbf{99.6}& 98.1& 10.0& 10.0& 95.9& 95.8& 85.3 &98.8\\
        CMUsubject1 & 60.0& \textbf{100.0}& 99.7& 98.3& 53.1& 51.0& 51.4& 97.6& 89.3&\textbf{ 100.0}\\
        CharacterTrajectories & 96.9& \textbf{99.0}& \textbf{99.0}& 97.1& 5.4&  	6.7& 93.8& 96.0&	92.0& 97.0 \\
        ECG & 74.8&	87.2& 86.7& 87.2& 67.0& 67.0& 50.0& 	84.1& 73.7 &\textbf{91.0}\\
        JapaneseVowels &97.6& \textbf{99.3}& 99.2& 97.6& 9.2& 	23.8& 94.4& 95.6& 96.5 & 98.7\\
        KickvsPunch &61.0& 54.0& 51.0& 61.0& 54.0& 	50.0& 	56.0& 62.0& 67.0 & \textbf{90.0}\\
        Libras &78.0& 96.4& \textbf{95.4}&  78.3& 6.7& 6.7& 	65.1& 	63.7& 79.4 &88.9\\
        NetFlow & 55.0&  89.1&  62.7&  77.7&  77.9&  	72.3&  63.0&  89.0&  94.5 &\textbf{100.0}\\
        UWave &90.1&  \textbf{93.4}&  92.6&  90.8&  12.5&  12.5&  	84.5&  85.9&  75.4 & 91.0\\
        Wafer & 89.4&  98.2&  98.9&  98.6&  89.4&  	89.4&  	65.8&  	94.8&  94.9 &\textbf{99.1}\\
        WalkvsRun &70.0&  \textbf{100.0}&  \textbf{100.0}&  \textbf{100.0}&  	75.0&  	60.0&  45.0&  	\textbf{100.0}&  94.4 &\textbf{100.0}\\
        PEMS  &-  &- &-  &-  &-  &-  &-  &-  &-  &93.6 \\
        \bottomrule         
        \end{tabular}
        
\end{table*}
\end{savenotes}

\subsection{Embedding}
In the original Transformers, the tokens are projected to a embedding layer. As time series data is continuous, we simply change the embedding layer to fully connected layer. Instead of linear projection, we add a non-linear activation $tanh$.  Following \cite{vaswani2017attention},
the positional encoding is added with the non-linearly transformed time series data to encode the temporal information, as self-attention is hard to utilize the sequential correlation of time step.

\subsection{Two-tower Transformer}
Multivariate time series has multiple channels where each channel is a univariate time series. The common assumption is that there exists hidden correlation between different channels at current or warping time step. Capturing both the step-wise (temporal) and channel-wise (spatial) information is the key for multivariate time series research. One common approach is to exploit convolutions. That is, the reception field integrates both step-wise and channel-wise by the 2D kernels or the 1D kernels with fixed parameter sharing. Different from other works that leverage the original Transformer for time series classification and forecasting, we designed a simple extension of the two-tower framework, where the encoders in each tower explicitly capture the step-wise and channel-wise correlation by attention and masking, as shown in Figure \ref{fig:GTN}.

\textbf{Step-wise Encoder}. To encode the temporal feature, we use the self-attention with mask to attend on each point cross all the channels by calculating the pair-wise attention weights among all the time steps. In the multi-head self-attention layers, the scaled dot-product attention formulates the attention matrix on all time step. Like the original Transformer architecture, position-wise fully connected feed-forward layers is stacked upon each multi-head attention layers for the enhanced feature extraction. The residual connection around each of the two sub-layers is also kept to direct information and gradient flow, following by the layer normalization.

\textbf{Channel-wise Encoder.} Likewise, the Channel-wise encoder calculates the attention weights among different channels across all the time step. Note that position of channel in the multivariate time series has no relative or absolute correlation, as if we switch the order of channels, the time series should has no changes. Therefore, we only add positional encoding in the Step-wise Encoder. Attention layers with the masking on all the channels is expected to explicitly capture the correlation among channels across all time step. Note that it is pretty straight-forward to implement both encoders by simply transpose the channel and time axis when feeding the time series for each encoders.

\begin{table*}[htbp]      
\centering
\caption{Ablation study of the two-towers, gating and masking in GTN}
\label{ablaton}
    \begin{tabular}{cccccc|c}    
        \toprule            
        & step-wise & step-wise+mask & channel-wise & channel-wise+mask & step-wise+channel-wise & GTN \\
        & &  &  & &+concatenation  & \\
        \midrule            
        AUSLAN &97.0& 95.3 & 94.5 & 94.5 & 96.7 & \textbf{97.5}\\       
        ArabicDigits & 98.6 & 98.5 & 98.3 & 98.3 & 	\textbf{98.9}& 98.8 \\
        CMUsubject16 &96.0 & 96.6 &\textbf{100.0} & \textbf{100.0} & 96.3 & \textbf{100.0}\\
        CharacterTrajectories & 96.0 & 97.1 & 96.9 & 96.5 & \textbf{97.5} & 97.0\\
        ECG & 87.0 & 84.0 & \textbf{92.0} & 89.0 & 86.0 &91.0\\
        JapaneseVowels & 96.0 & 97.3 & 98.3 & 97.6 & 98.1 & \textbf{98.7}\\
        KickvsPunch &81.3 & 80.0 & 81.3 &\textbf{ 90.0} & 81.3 & \textbf{90.0}\\
        Libras & 82.0 & 81.1 & 88.3 & 88.3 & \textbf{90.5} & 88.9\\
        NetFlow & 88.0 & \textbf{100.0} & \textbf{100.0} & \textbf{100.0} & \textbf{100.0} & \textbf{100.0}\\
        UWave &89.0 & 88.8 & 90.0 & 88.3 & 89.5 &\textbf{91.0}\\
        Wafer &97.0 & 98.1 & 96.9 & 97.9 & 97.9 & \textbf{99.1}\\
        WalkvsRun &96.4 & \textbf{100.0} & 96.5 & \textbf{100.0} & \textbf{100.0} & \textbf{100.0}\\
        PEMS & \textbf{94.0} & 92.5 & 87.9 & 91.4& 90.8 &93.6\\
        \bottomrule         
        \end{tabular}
\end{table*}

\subsection{Gating}
To merge the feature of the two towers which encodes step-wise and channel-wise correlations, a simple way is to concatenate all the features from two towers, which compromises the performance of both as shown in our ablation study.

Instead, we proposed a simple gating mechanism to learn the weight of each tower. After getting the output of each tower, which had a fully connect layer after each output of both encoders with non-linear activation as $C$ and $S$, we packed them into vector by concatenation followed by a linear projection layer to get $h$. After the softmax function, the gating weight are computed as $g_1$ and $g_2$. Then each gating weight is attending on the corresponding tower's output and packed as the final feature vector.

\begin{eqnarray}
h = {\bf{W}} \cdot Concat(C, S) + b \nonumber \\
g_1, g_2 = Softmax(h) \nonumber \\
y = Concat(C \cdot g_1, S \cdot g_2) \nonumber \\
\label{eqn:gating}
\end{eqnarray}

\section{Experiments}
\subsection{Experiment Settings}
We test GTN on the same subset of the Baydogan archive \cite{baydogan2019multivariate}, which contains 13 multivariate time series datasets. All the datasets have been split into training and testing by default, and there is no preprocessing for these time series. We choose the following deep learning models as the benchmarks. 

\begin{itemize}
\item  Fully Convolutional Networks (\textbf{FCN}) and Residual Networks (\textbf{ResNet}) \cite{7966039}. These are reported to be among the best deep learning models in the multivariate time series classification task \cite{IsmailFawaz2018deep}. Multi-layer Perception (\textbf{MLP}) is also included as a simple baseline in our comparison.
\item Universal Neural Network Encoder (\textbf{Encoder}) \cite{serra2018towards}.
\item Multi-scale Convolutional Neural Network (\textbf{MCNN}) \cite{cui2016multi}.
\item  Multi Channel Deep Convolutional Neural Network (\textbf{MCDCNN}) \cite{zheng2016exploiting}.
\item  Time Convolutional Neural Network (\textbf{Time-CNN}) \cite{zhao2017convolutional}.
\item Time Le-Net (\textbf{t-LeNet}) \cite{le2016data}.
\item Time Warping Invariant Echo State Network (\textbf{TWIESN}) \cite{tanisaro2016time}.
\end{itemize}

The Gated Transformer Network is trained with Adagrad with learning rate 0.0001 and dropout = 0.2. The categorical cross-entropy is used as the loss function. Learning rate schedule on plateau \cite{7966039,IsmailFawaz2018deep} is applied to train the GTN. We test on the training set and the test set every certain number of iterations, the best test results and its super parameters would be recorded. For fair comparison, we choose to report the test accuracy on the the model with the best training loss as \cite{IsmailFawaz2018deep}.\footnote{The codes are available at https://github.com/ZZUFaceBookDL/GTN.
}

\subsection{Results and Analysis}

The results are shown in Table \ref{results}. GTN achieved comparable results with the FCN and ResNet. Note that the results has no statistical significant difference among these three models, though on \textit{NetFlow} and \textit{KickvsPunch} datasets, GTN shows superior performance. The drawback of GTN is comparably leaning to overfitting. Unlike FCN and ResNet where no dropout is used, the GTN has dropout binding with layer norm to reduce the risk of overfitting. 

\subsubsection{Ablation Study}
To clearly state the performance gain from each module in the GTN, we performed a comprehensive study as shown in Table \ref{ablaton}. 
\begin{itemize}
\item Following the traditional Transformer, masking helps to ensure that the predictions for position $i$ can depend only on the known previous outputs and also helps the attention to not attend to the padding position. This benefits holds not only on the language but also on the time series data, as the tower-only transformer with mask are overall a bit better than the one without masks.  
\item Channel-wise only transformer outperforms step-wise only transformer on the majority of the dataset. This preliminary result supports our assumption that for multivariate time series, the correlation between different channels across all time step is an important differentiator with the univariate time series. The attention with mask is able to catch the channel-wise feature better.
\item Different time series data might show different leaning on channel-wise and step-wise information. For example, on the dataset \textit{PEMS}, the step-wise Transformer model works better. On the dataset \textit{CMUsubject16}, the channel-wise Transformer model outperforms. 
\item Following the above point, exploiting both towers is a straight-forward solution. However, simple concatenation of the two towers sometimes works, but the performance always fall on the middle ground or even worse. Like on the dataset \textit{PEMS} and \textit{CMUsubject16}, step-wise model and channel-wise model works best for each cases. After concatenation of both towers' feature, the results becomes even worse. 
\item By adding the gating weights before simple (equally-weighted) concatenation, the model is able to learn when to rely on a specific tower more by a pure data-driven way, thus show the best performance in this study.
\end{itemize}

\subsection{Visualization and Analysis of the Attention Map}

\begin{figure}
    \centering
    \includegraphics[scale=0.4]{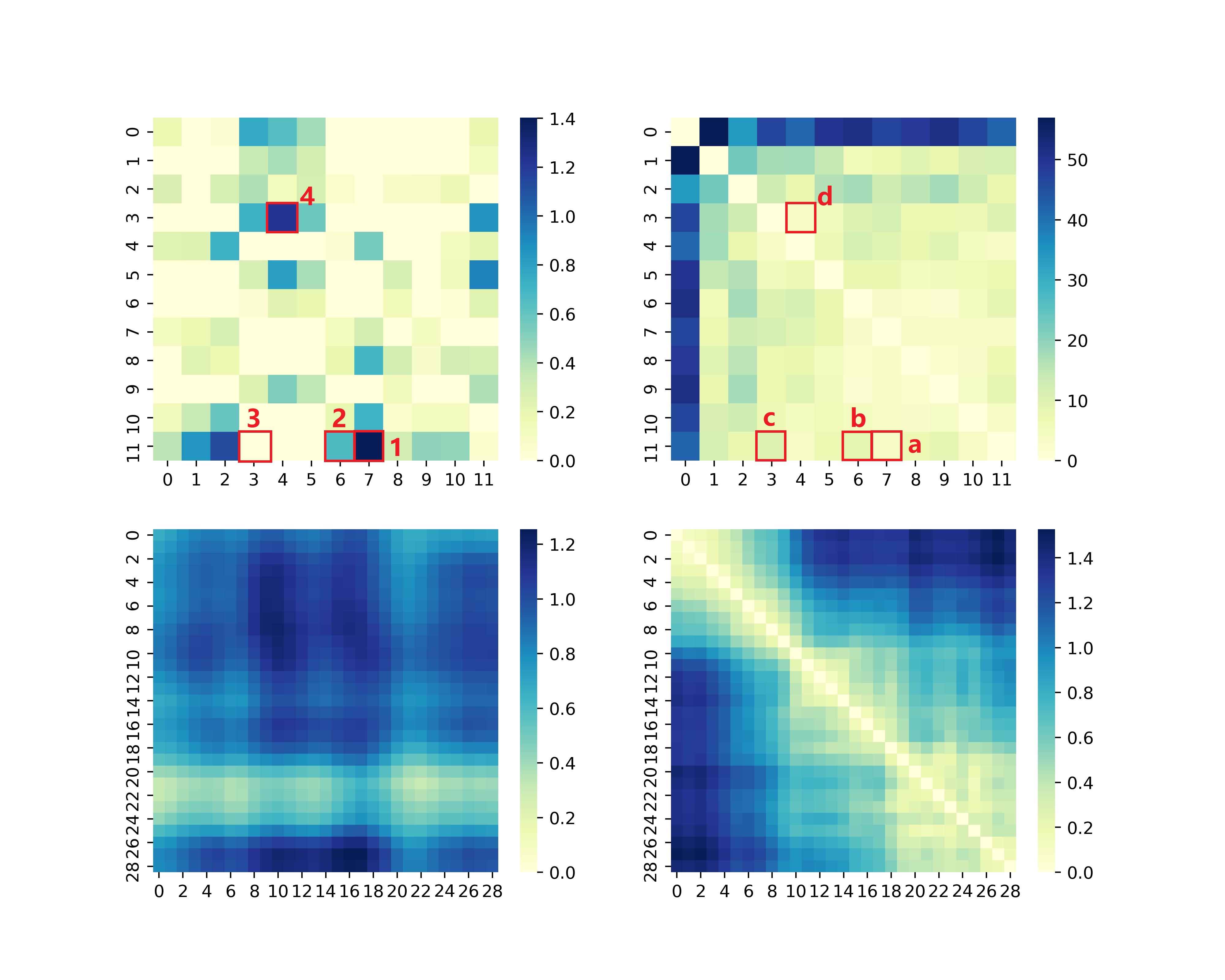}
    \caption{1) channel-wise attention map (upper-left) 2) channel-wise DTW (upper-right) 3) step-wise attention map (bottom-left) 4) step-wise L2 distance (bottom-right)}
    \label{fig:my_label_attention}
\end{figure}

The attention matrix represents the correlation between  the channels and time steps respectively. We choose one sample from the \textit{JapaneseVowels} dataset to visualize both attention maps. for the channel-wise attention map, we calculated the dynamic time warping (DTW) distance across the time series on different channel. For each time step, we also simply calculated the Euclidean distance across different channels, as on the same time step there is no time axis, thus DTW is not needed. The visualization is shown in Figure \ref{fig:my_label_attention}

\begin{figure}
    \centering
    \includegraphics[scale=0.23]{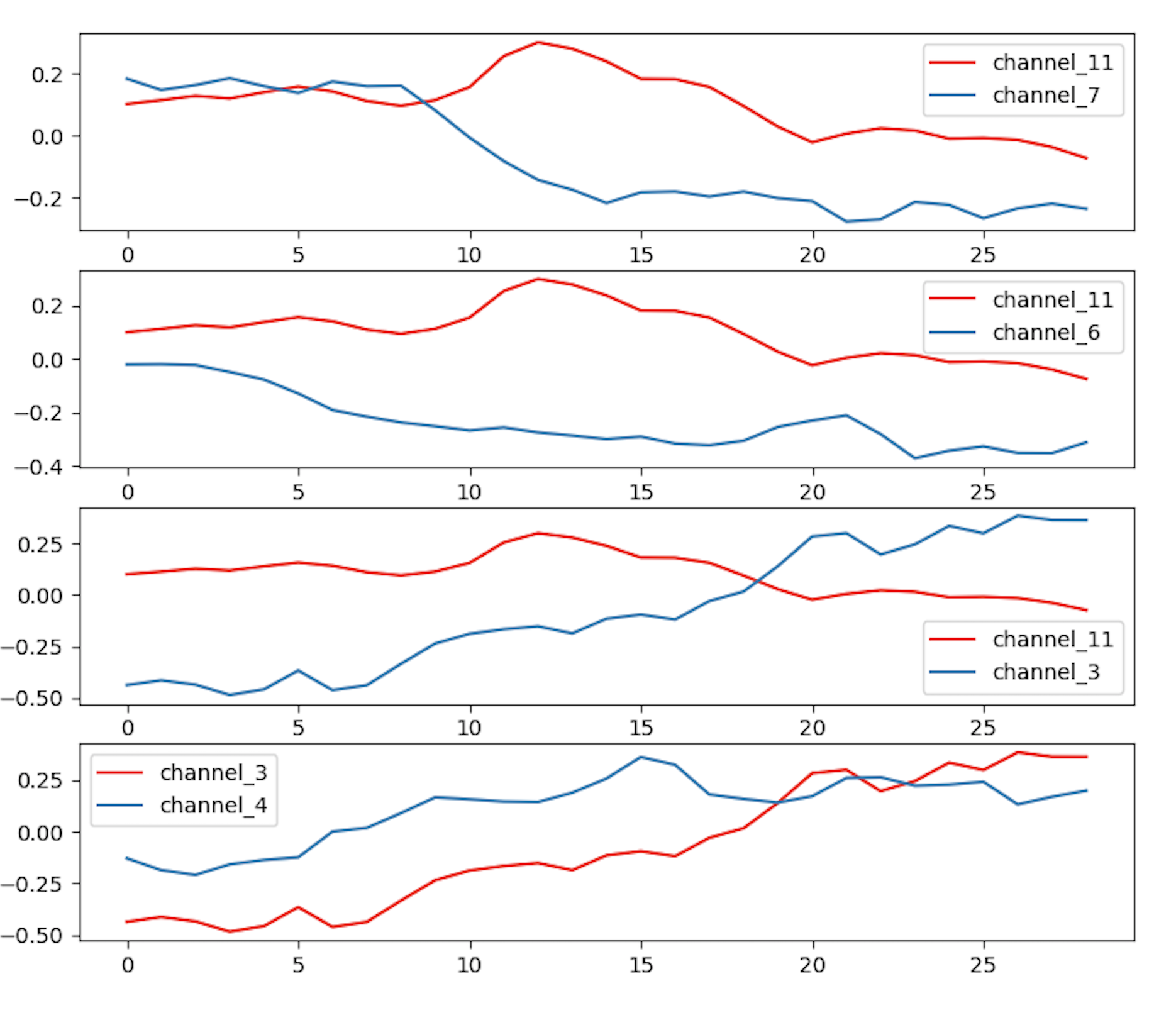}
    \caption{Drawing of the raw time series in different channel pairs.}
    \label{fig:feature_compare}
\end{figure}

Our first analysis focuses on the channel-wise attention map. In Fig \ref{fig:my_label_attention}, we label four blocks ($b1$-$b4$). In Fig \ref{fig:feature_compare}, we draw the raw time series of the corresponding channels. We use $channel_{11}$ (abbreviated as $c11$, the same below) to represent channel 11 in the following analysis. 

$b1$ and $b2$ have relatively high attention score, that is, both $c6$ and $c7$ are strongly co-fired with $c11$. Look at Fig \ref{fig:feature_compare}, we can clearly see that $c6$ and $c7$ have relatively similar shapelet and trend with $c11$. Similarly, $b4$ have very high attention score, where $c3$ and $c4$ are also showing very similar trend. Meanwhile, $b3$ is among the one with very small attention score. Look at the $c11$ and $c3$, two time series show pretty much different even inverse trend. This preliminary finding is interesting because like the attention in NLP task where the attention score indicates the semantic correlation between different tokens, the channel-wise attention learned from  the time series similarly shows the similar sequences that are co-fired together to force the learning lean to the final labels.

Note that the smaller DTW does not mean two sequence are similar. Like shown on the channel-wise DTW, block $c$ indicates very small DTW distance between $c3$ and $c11$. Actually $c11$ and $c3$ are very different in trends and shapelet. Our preliminary analysis shows that the channel-wise attention also tends to grab the similar sequences where DTW shows no clear differentiated factors.

As the step is fixed across channel for the step-wise attention, Euclidean distance is the special case of DTW at the same step, thus we choose Euclidean distance as the metric. Through the analysis of step-wise attention map and the Euclidean distance matrix, the distance between the time series and the similarity of the shapelet have an impact on who GTN calculates the attention scores, though the impact are not that obvious which deserves a deep dive in the future work.

\subsection{Analysis on the Gating Weight}
GTN has two towers to encode the step-wise and channel-wise information with self-attention respectively. With the gating method, the model learns to assign different weights to attend to these two Transformers. In the ablation study, we show that gating achieved better results compared with concatenation. 

By observing the gate weights assigned to the two tower on the \textit{AULSAN} data set, we found that for different sample, the gating weights assigned to step-wise and channel-wise tower are tends to be different, like $\{0.9899, 0.0101\}$ and $\{0.0443, 0.5557\}$ for two sample time series respectively. However, the gating weights overall show the trends to be skewed towards the step-wise tower, with the average gating weights of $\{0.7786, 0.2214\}$. As shown in Table \ref{ablaton}, the step-wise Transformer outperforms the channel-wise Transformer, thus overall the gating learns to assign more weights to the step-wise tower, the gating behavior and the results in the ablation study are consistent. The gating shows the capability to learn the weights from the data-driven manner to attend on each towers for different samples and dataset to improve the classification performance.

\subsection{Analysis of the Embedding Output}
\begin{figure}
    \centering
    \includegraphics[scale=0.5]{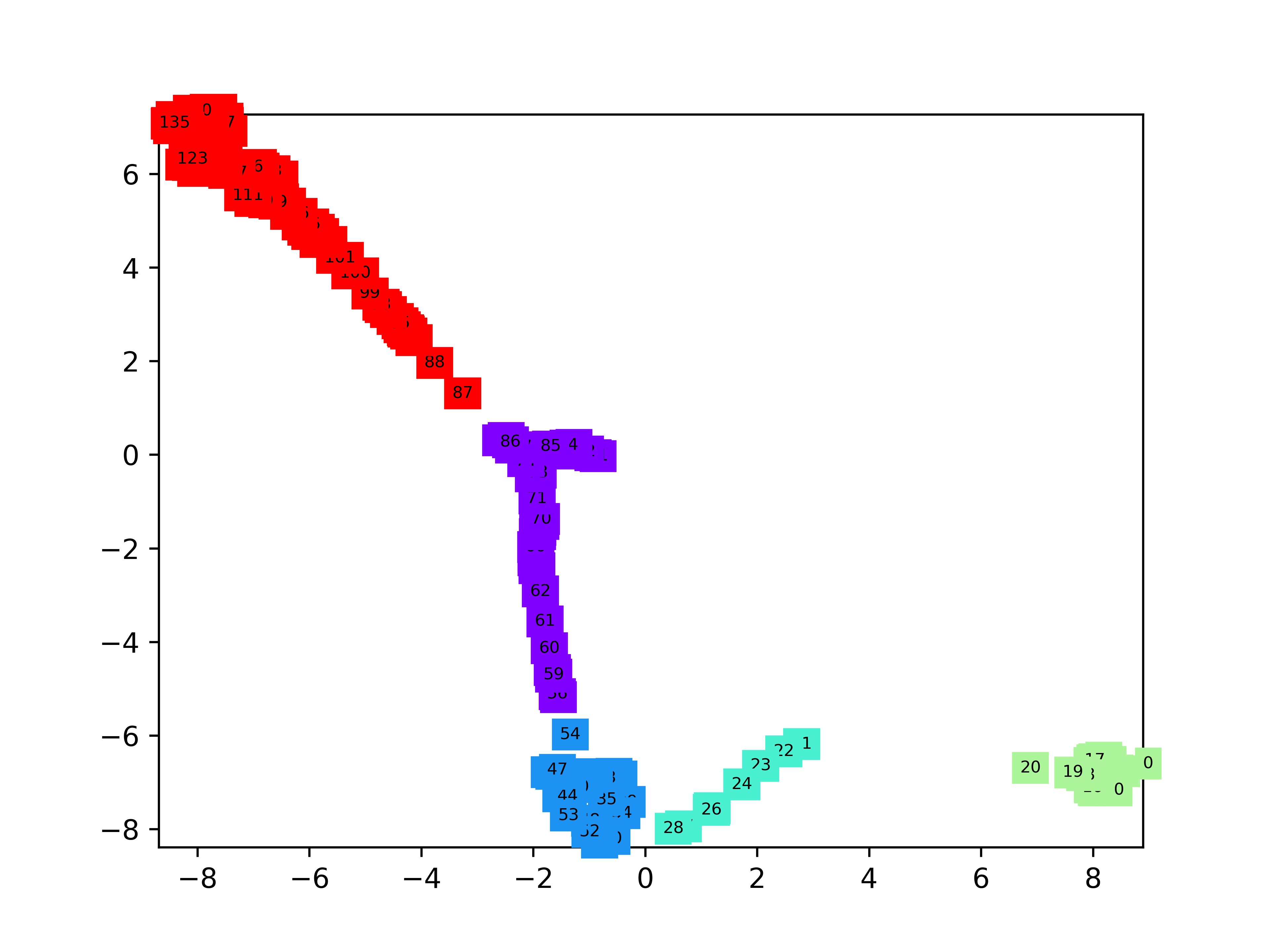}
    \caption{Visualization of the t-SNE result of the embedding layer output on the \textit{AUSLAN} dataset.}
    \label{graph:tsne}
\end{figure}

As each time step are transformed to a dense vector through the embedding layer, we use the t-SNE \cite{maaten2008visualizing} to reduce the dimension of the output from the embedding layer on \textit{AUSLAN}. The result is shown in graph \ref{graph:tsne}. The graph shows that all the points are clustered together on the specific manifold. We roughly labeled those clusters with five different colors.

\begin{figure}
    \centering
    \includegraphics[scale=0.45]{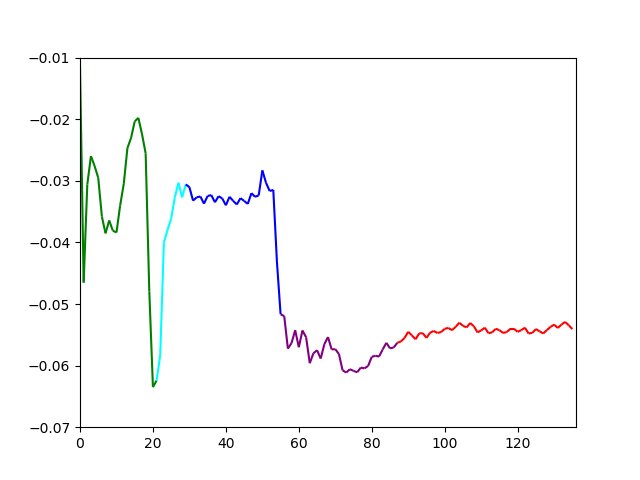}
    \caption{Shapelet discovered by the visualization of the embedding on each time step.}
    \label{graph:rawtimeseries}
\end{figure}

As shown in Figure \ref{graph:rawtimeseries}, we mapped each color back to the raw time series. Interestingly, the time step from each cluster are consistent and overall each cluster shows different shapelet. 

\begin{itemize}
\item The green shapelet indicates a deep \textit{M} shape.
\item The light blue shapelet is a sharp trending-up sub-sequence.
\item The dark blue shapelet is a plateau followed by a deep down like $7$.
\item The magenta shapelet is like a recovering from the deep down with a bit trending up.
\item The red shapelet is a plateau.

By a closer look at the visualization of the embedding on each time step, the near points that forms some interesting shapelet has relatively smaller distance in the vector space. Like \cite{liu2016encoding}, where the author discover interesting shapelet by a well designed Markov transition matrix with the clustering on the transformed complex network graph, GTN shows the potential by learning the interesting pattern of some specific sub-sequences from scratch, which will be an interesting future work.

\end{itemize}

\subsection{Visualization of the Extracted Feature}
\begin{figure}
    \centering
    \includegraphics[scale=0.6]{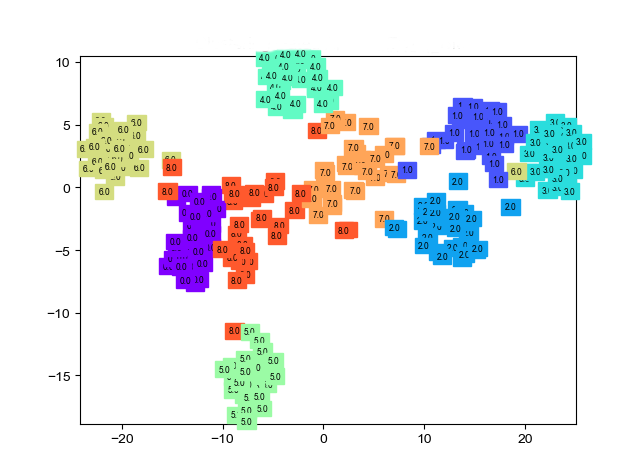}
    \caption{Visualization of the feature extracted after the 
    gating in GTN.}
    \label{graph:clster}
\end{figure}
lastly, following \cite{IsmailFawaz2018deep}, we visualize the 
feature vector after gating for a simple sanity check of the 
feature quality. We choose the \textit{JapaneseVowels} dataset 
and apply t-SNE to reduce the dimension for visualization, as 
shown in Figure \ref{graph:clster}. Each number in the graph 
is the label with corresponding color encoding. and labels in 
the figure correspond to the labels of the original data. This 
figure shows that GTN is able to project the data into an 
easily separable space for better classification results.
 
\section{Conclusion}
We presented the Gated Transformer Network (GTN) as a simple 
extension of multidimensional time series using gating. With 
the gating that merges two towers of transformer networks 
which model the channel-wise and step-wise correlations 
respectively GTN is able to explicitly learn both channel-wise 
and step-wise correlations. We conducted comprehensive 
experiments on thirteen dataset and the preliminary results 
show that GTN is able to achieve competing performance with 
current state-of-the-art deep learning models. The ablation 
study shows how different modules work to together to achieve 
the improved performance. We also qualitatively analyzed the 
attention map and other components with visualization to 
better understand the interpretability of out model. Our 
preliminary results ground a solid foundation for the study of 
Transformer Network on the time series classification task in 
future research.

\newpage
\bibliographystyle{named}
\bibliography{ijcai21}

\end{document}